\documentclass[10pt,letterpaper]{article}
\usepackage[top=0.85in,footskip=0.75in]{geometry}

\usepackage{amsmath,amssymb}

\usepackage{changepage}

\usepackage[utf8x]{inputenc}

\usepackage{textcomp,marvosym}

\usepackage{cite}

\usepackage{nameref,hyperref}

\usepackage{microtype}
\DisableLigatures[f]{encoding = *, family = * }

\usepackage[table]{xcolor}

\usepackage{array}

\newcolumntype{+}{!{\vrule width 2pt}}

\newlength\savedwidth

\newcommand\thickhline{\noalign{\global\savedwidth\arrayrulewidth\global\arrayrulewidth 2pt}%
\hline
\noalign{\global\arrayrulewidth\savedwidth}}

\usepackage[aboveskip=1pt,labelfont=bf,labelsep=period,justification=raggedright,singlelinecheck=off]{caption}

\makeatletter
\renewcommand{\@biblabel}[1]{\quad#1.}
\makeatother

\date{}

\usepackage{lastpage,fancyhdr,graphicx}
\usepackage{epstopdf}
\usepackage{algorithm}
\usepackage{algpseudocode}
\usepackage{pifont}
\usepackage{subcaption}
\usepackage{multirow}
\usepackage{tabularx}
\usepackage{bigstrut}

\newcolumntype{Y}{>{\centering\arraybackslash}X}

\usepackage{graphicx}

\begin{document}
\vspace*{0.2in}

\begin{flushleft}
{\Large
\textbf\newline{Deep Learning the Indus Script} 
}
\newline
\\
Satish Palaniappan\textsuperscript{1*},
Ronojoy Adhikari\textsuperscript{2}
\\
\bigskip
\textbf{1} Department of Computer Science and Engineering, Sri Sivasubramaniya Nadar College of Engineering, Chennai, India - 603110
\\
\textbf{2} The Institute of Mathematical Sciences, Chennai, India - 600113
\\
\bigskip

%
%
* E-mail: tpsatish95@gmail.com, satish12095@cse.ssn.edu.in

\end{flushleft}
\section*{Abstract}
Standardized corpora of undeciphered scripts, a necessary starting point for computational epigraphy, requires laborious human effort for their preparation from raw archaeological records. Automating this process through machine learning algorithms can be of significant aid to epigraphical research. Here, we take the first steps in this direction and present a deep learning pipeline that takes as input images of the undeciphered Indus script, as found in archaeological artifacts, and returns as output a string of graphemes, suitable for inclusion in a standard corpus. The image is first decomposed into regions using Selective Search and these regions are classified as containing textual and/or graphical information using a convolutional neural network. Regions classified as potentially containing text are hierarchically merged and trimmed to remove non-textual information. The remaining textual part of the image is segmented using standard image processing techniques to isolate individual graphemes. This set is finally passed to a second convolutional neural network to classify the graphemes, based on a standard corpus. The classifier can identify the presence or absence of the most frequent Indus grapheme, the ``jar” sign, with an accuracy of 92\%. Our results demonstrate the great potential of deep learning approaches in computational epigraphy and, more generally, in the digital humanities.

\section*{Introduction}
Epigraphical research is substantially facilitated by the availability of standardized corpus of inscriptions. These must be collated from their original context in the archaeological record, most often from artifacts of stone or clay, and mapped to the standard set of graphemes in the corpus. Morphosyntactic and semantic studies of the inscriptions can then be based largely on the corpus, with references to the original context being rarely necessary. The task of compiling a corpus from the raw archaeological record, however, requires a great deal of care and expertise and it is one of the most laborious aspects of epigraphical research.

Automation via computer programs has been used in the past \cite{knorozov1968proto} \cite{koskenniemi1970method} \cite{koskenniemi1981syntactic} \cite{rao2009entropic} \cite{rao2009markov} to reduce human effort in epigraphical research. However, without exception, these have been confined to classification \cite{koskenniemi1970method} and search for graphemic patterns \cite{koskenniemi1981syntactic} \cite{rao2009entropic} \cite{rao2009markov} in a prepared corpus. Automation in the complementary task of corpus preparation has, to the best of our knowledge, not been attempted yet.

Here, we take the first steps towards automated corpus compilation by constructing a deep learning \cite{deep_learning2015}\cite{deep_learning_conspiracy2015} pipeline that takes as input, images of archaeological artifacts, and returns as output, a string of graphemes, suitable for inclusion in a standard corpus. We use images of archaeological records of the Indus script \cite{kenoyer1998ancient}, mostly found inscribed on terracotta seals, to demonstrate the efficacy of the pipeline.

The deep learning pipeline consists of two principal components. The first is a convolutional neural network \cite{cnn_lenet_lecun1998gradient} that identifies three types of regions in the image, those that contain a sequence of graphemes (henceforth abbreviated as ``text"), those that contain images or other non-graphemic components (abbreviated as ``no-text") and those that contain both graphemic and non-graphemic components (abbreviated as ``both"). The output of this classifier is then received by a second convolutional neural network \cite{cnn_lenet_lecun1998gradient} which identifies graphemes from the textual region of the image and maps then to a standard set of graphemes. In the present work, this second stage has been implemented to recognize only the most frequent Indus grapheme, the ``jar sign" (sign 342 in the Mahadevan corpus \cite{mahadevan1977indus}). There are several intermediate image processing components that are described in detail below. The code and other resources used in constructing this deep learning pipeline has been open sourced and is available on GitHub \cite{ocr_indus_seals2016}.

Our work demonstrates the potential of cutting edge deep learning algorithms in transforming humanities research, including the study of language and literature.

\section*{Results and Discussion}
\begin{figure}[!ht]
\centering
\includegraphics[width=\textwidth]{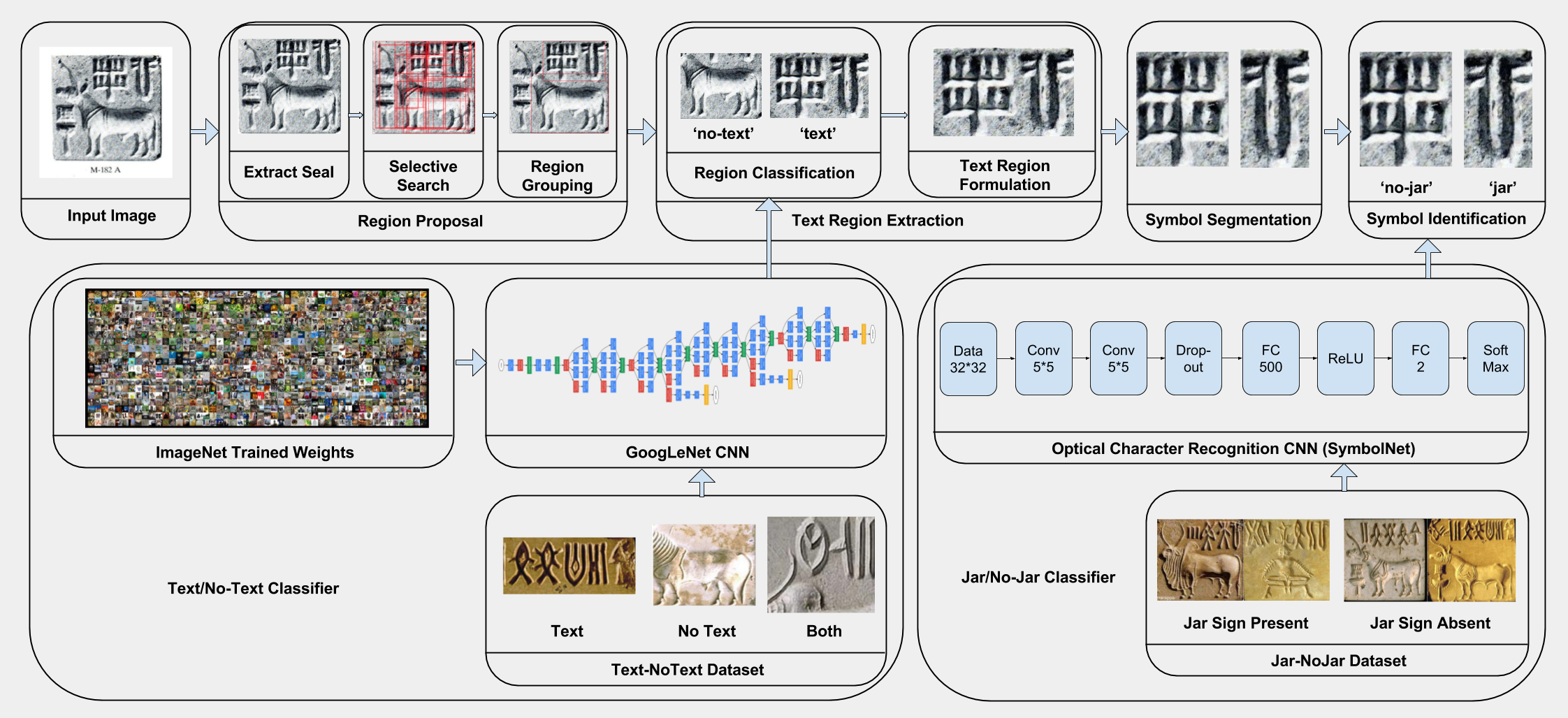}
\caption{{\bf The proposed deep learning pipeline}}
\label{pipeline}
\end{figure}

Our principal result is in Fig~\ref{pipeline}, where the ``flow" of the image of an Indus seal is shown as it progresses through the pipeline. A typical seal, shown at the top left of Fig~\ref{pipeline}, contains a region of text (usually at the top of the seal) and non-textual iconographic elements (usually below the text). This image is input into the deep learning pipeline and the final output, shown at the top right, is the sequence of graphemes contained in the textual region of the image, together with a classification of each grapheme. At present, we classify graphemes as belong to the ``jar" class or to its complement. A complete classifier will be implemented in future. We now describe the various components of the pipeline, with specific detail to those components shown in the lower part of Fig~\ref{pipeline}.

\subsection*{Region Proposal}
\begin{figure}[!ht]
\centering
\includegraphics[width=\textwidth]{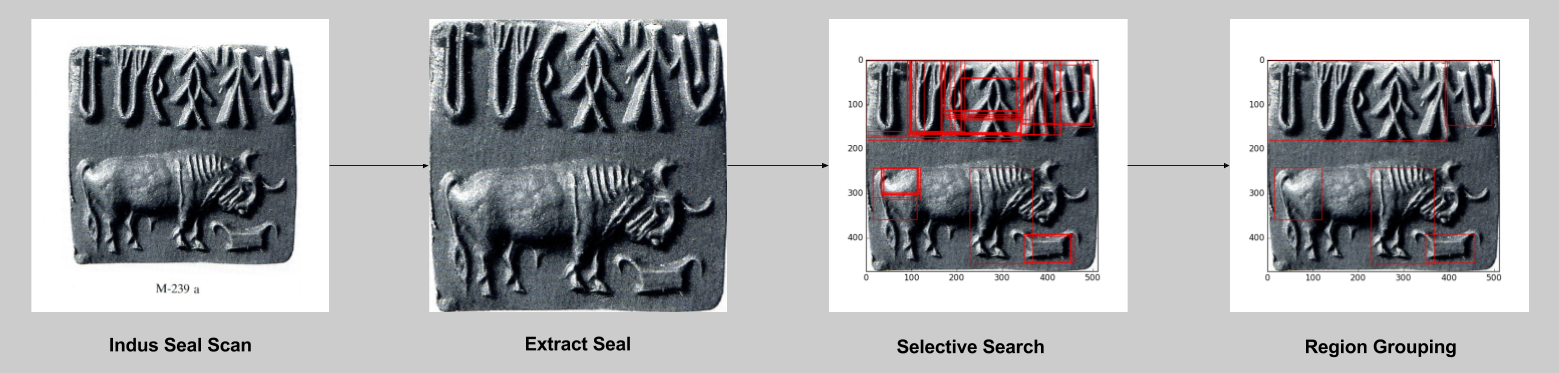}
\caption{{\bf The Region Proposal stage}}
\label{region_proposal}
\end{figure}

The first stage of the pipeline proposes regions of interest that have a high probability of containing a symbol, animal, deity or any iconographic element that is inscribed on the Indus seals. This stage has three sub-stages namely, Extract Seal, Selective Search, and Region Grouping for progressively extracting the regions of interest. Fig~\ref{region_proposal} is a sample flow illustrating this stage of the pipeline.

\subsubsection*{Extract Seal}
This stage takes the image as input and extracts the seal portion only by removing the irrelevant background information from the image using standard image processing techniques. The input image is first grayscaled and smoothed using a multi-dimensional Gaussian filter with a kernel of standard deviation $3.0$, as the Indus seals are characterized by heavy wear and tear that leads to noisy images. Then each pixel is thresholded at the seal's background mean pixel value following another level of smoothing with the Gaussian blur of $7x7$ kernel size. This repeated smoothing ensures that only the most prominent edges constituting the entire seal will be detected when we perform an optimized canny edge \cite{canny1986computational} detection (see Materials and Methods) to propose contours for calculating the bounding rectangular box around the seal.

\subsubsection*{Selective Search}
These cropped seal images are taken as the input to the next stage that proposes all possible regions of interest that are likely to have the Indus script symbols or depictions of animals like the bull, unicorn, deities, and other iconographies. We use the Selective Search \cite{uijlings2013selective} (see Materials and Methods) region proposal algorithm to achieve this task. However, the vanilla version of this algorithm had to be fine tuned by grid searching over the four free parameters to better suit our use case of ultimately extracting text only regions.

\subsubsection*{Region Grouping}
In order to reduce the numbers and increase the quality of the proposed regions, a four-level hierarchy of region grouping methodology was devised to improvise the Selective Search \cite{uijlings2013selective} results. At the first \textbf{merge concentric proposals} level, all those proposals that were mere approximations and generalizations of each other, thresholded at 14\% of the arithmetic mean of the proposals' dimensions, were replaced with a single enclosing region of dimensions equal to the mean of all concentric proposals. Secondly, the \textbf{contained boxes removal} level removes those proposals that were 100\% encompassed inside another region proposal, thus only considering the region proposal covering whole symbols or objects rather than each part of it. Further, the \textbf{draw super box} level replaced all the proposals that were overlapping with over 40\% of each other's area, such that a supposed unified region was proposed as two different yet partially overlapping regions, with a single minimal super box that bounded both the proposals. Finally, the \textbf{draw extended super box} level replaces those regions in hand, that were continuous regions in the seal image arranged along the horizontal or vertical axes of the image, offset only by 6\% of the arithmetic mean of the proposals' dimensions, with a single horizontal/vertical super region. Effectively, the last two levels leverage the prior information that the text regions are contiguous, being mostly arranged along a same line or axis rather than randomly distributed in space. Additionally, the above replace, remove or merge operations over the region proposals are decided based on thresholds determined by empirical search.

\subsection*{Text Region Extraction}
\begin{figure}[!ht]
\centering
\includegraphics[width=\textwidth]{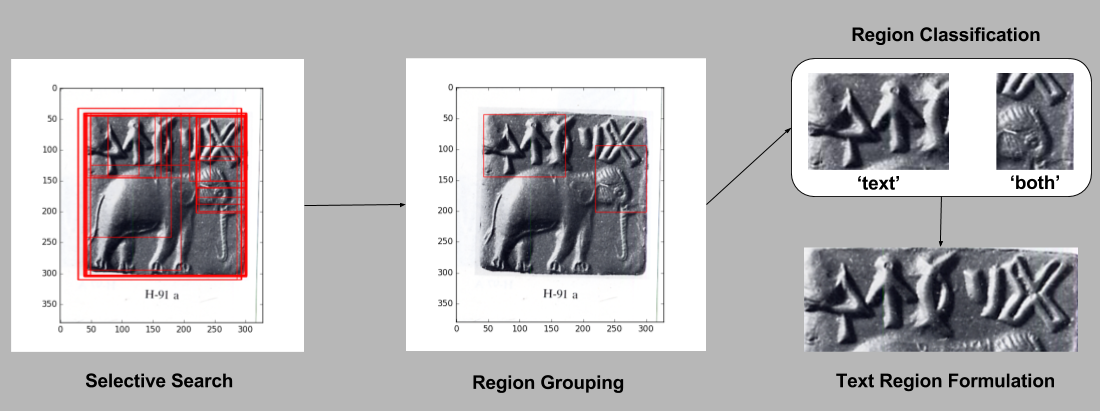}
\caption{{\bf The Text Region Extraction stage}}
\label{text_region_extraction}
\end{figure}

This stage of the pipeline takes the candidate region of interest proposals from the previous stage as input and produces precise regions that contain only textual information by eliminating the non-symbolic parts off of the region proposals. To achieve this, we have two sequential sub-stages, Region Classification and Text Region Formulation. Fig~\ref{text_region_extraction} illustrates a sample flow of this stage of the pipeline.

\subsubsection*{Region Classification}
This sub-stage of the pipeline receives the region of interest proposals from the Region Proposal stage as input and classifies them into three types namely, ``text", ``no-text", and ``both" regions. The ``text" and ``no-text" regions consist, respectively, of only the Indus script graphemes and only the non-graphemic elements. The ``both" regions consist of both script and non-script portions in it.

In order to achieve this, we trained a machine learned model, the Text/No-Text classifier, that can differentiate a text portion from a non-text portion, in any given input seal region of interest. This is ultimately an image classification problem with three classes namely, ``text", ``no-text", and ``both". In order to clearly differentiate between these classes, we extract deep features that adapt themselves by hierarchically learning what and where to focus on in the input images given the training samples, instead of hand-crafting it. Especially, the images of the Indus seals, that come from archaeological sites are just remains of the great civilization, that is mostly broken, scratched, and worn out. In addition to this, the different elevations to the text added by the seals and sealings, partially erased seals, non-uniformity in scale, text size and spacing, dynamic lighting conditions, similarity between symbols, and above all relatively small dataset in the range of only few thousand images, are all the problems that must be solved to train this classifier. Thus, the deep feature extraction technique proved to be a perfect fit by being robust to the cross product of symbol scale variations, photograph or scan illumination variations, deformations due to serious wear and tear, photograph with background clutter and symbol intra-class variations that the input Indus seal regions of interest suffer.

\begin{figure}[!ht]
\centering
\includegraphics[width=\textwidth]{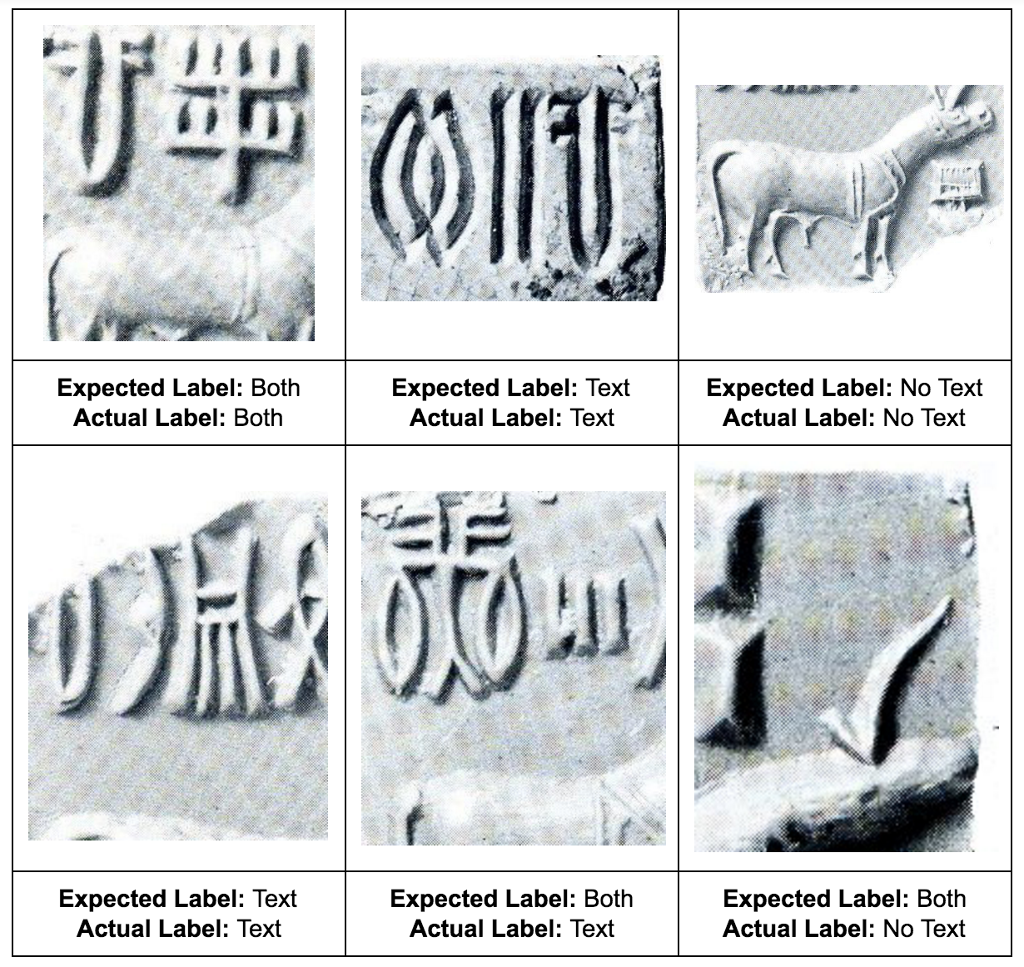}
\caption{{\bf The Region Classification sub-stage's Text/No-Text CNN's sample classification outcomes}}
\label{region_classification}
\end{figure}

Therefore, a Convolutional Neural Network (CNN) with a terminal SoftMax classification layer was trained over the ``Text-NoText Dataset" (see Materials and Methods). The lightweight GoogLeNet (see Materials and Methods), which is 22 layers deep with 9 inception modules (network within a network) was used as the base CNN architecture for this purpose. We opted for applying transfer-learning and fine-tuning approaches over the GoogLeNet CNN, initialized with extremely feature-rich filters (weights) pre-trained over the ImageNet dataset \cite{ImageNet_russakovsky2015imagenet}, given the need for developing a highly discriminative classifier with a comparatively very small and unconventional base dataset consisting of just 2091 data samples. All the inception layers except the last three were frozen and made devoid of any further learning, while all the last three inception layers' default learning rates were doubled. This freezes the rich and generic lower level non-dataset specific ImageNet weights, while actively learning the Indus dataset specific higher level features. Finally, the last fully-connected layer is rewired to output just three values each for one destination class in our current use case of Indus text region classification, which is then classified by the terminal SoftMax classifier in the CNN. Fig~\ref{region_classification} exhibits a sample of the classification results from this sub-stage of the pipeline.

\subsubsection*{Text Region Formulation}
This sub-stage of the pipeline takes the labeled region proposals from the previous sub-stage as input and isolates the text-only regions from them. In order to achieve this, we have formulated a two-level hierarchy to clip off the non-text regions and to group all the text regions into one single strip. Firstly, the \textbf{Draw TextBox} level merges those pairs of region proposals, where two ``text" regions or a ``text" region and a ``both" region are aligned along the same horizontal or vertical axis within a threshold difference both in terms of height and width, irrespective of whether they are overlapping or not, in order to get whole text regions into one single proposal called ``text box"; the dynamic threshold over the merge operation, is calculated by computing the arithmetic mean of the width and height of the two candidate regions and taking either 25\% or 20\% of it, respectively. Secondly, the \textbf{Trim TextBox} level is responsible for clipping off the non-textual information (``no-text") in those pairs of region proposals, where a ``text box"/``text" region and a ``no-text" region were overlapping, in order to get the perfect text-only regions. This trim operation happens if both of the candidate regions have an overlap of 70\% across the height and 20\% across the width or the vice versa. Both of these levels once again harnesses the fact that, most of the Indus seals have the symbols arranged either horizontally or vertically along an axis.

\subsection*{Symbol Segmentation}
This is the penultimate stage of the pipeline that takes the precise text-only region proposals as input and segments out the individual graphemes from them. For achieving this a customized algorithm that stacked together various standard image processing techniques was devised as outlined in Algorithm~\ref{segment_symbol}. This stage ultimately reduces the optical character recognition problem to a much simpler symbol classification problem, by extracting out the individual symbols.

\begin{algorithm}
\caption{Symbol Segmentation Algorithm}
\label{segment_symbol}
\begin{algorithmic}[1]
\Procedure{Segment\textendash Symbol}{Image I}
\State Gray\textunderscore Image = \textbf{Gray\textunderscore Scale}(I)
\State Thresholded\textunderscore I = \textbf{Otsu\textunderscore Thresholding}(Gray\textunderscore Image)
\State Smoothened\textunderscore Image = \textbf{Gaussian\textunderscore Blur}(Thresholded\textunderscore Image)
\State Component\textunderscore ROIs = \textbf{Connected\textunderscore Colour\textunderscore Components}(Smoothened\textunderscore Image)
\State ROIs = \textbf{Combine}(Component\textunderscore ROIs)
\State \hspace{\algorithmicindent} Unique\textunderscore ROIs = \textbf{Contained\textunderscore Boxes\textunderscore Removal}(Component\textunderscore ROIs)
\State \hspace{\algorithmicindent} Super\textunderscore ROIs = \textbf{Draw\textunderscore Super\textunderscore Box}(Unique\textunderscore ROIs)
\State \hspace{\algorithmicindent} ROIs = \textbf{Draw\textunderscore Extended\textunderscore Super\textunderscore Box}(Super\textunderscore ROIs)
\State Segmented\textunderscore Symbols = \textbf{Crop}(ROIs, I)
\EndProcedure
\end{algorithmic}
\end{algorithm}

\begin{figure}[!ht]
\centering
\includegraphics[width=\textwidth]{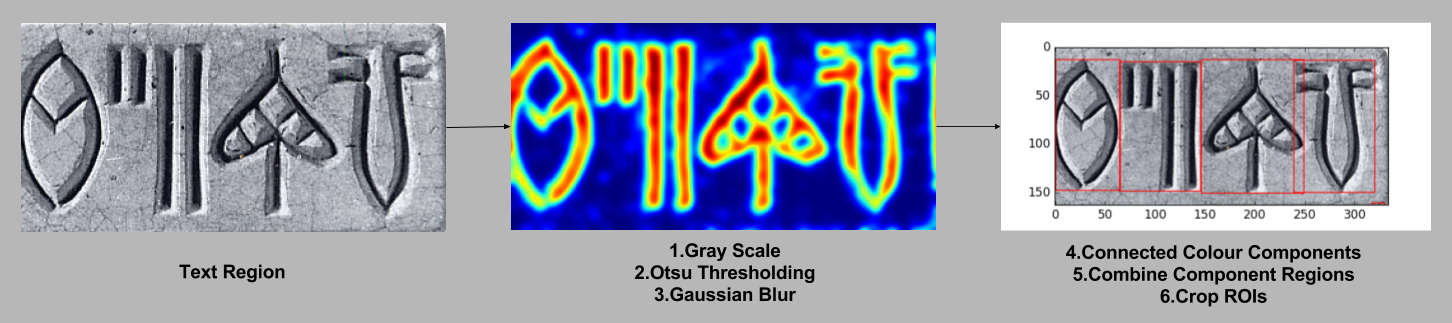}
\caption{{\bf The Symbol Segmentation stage}}
\label{symbol_segmentation}
\end{figure}

The stage uses the six level Algorithm~\ref{segment_symbol} for clipping out the individual symbols from the input text-only regions. The first level converts the input regions into grayscale. This grayscaled image is then binarized using Otsu's thresholding \cite{otsu1975threshold} technique in the second level; we settled on Otsu's thresholding technique for our purpose, after experimenting with various techniques like, Mean, Triangle, Li \cite{li1993minimum}, Yen \cite{yen1995new}, IsoData \cite{isodata_ridler1978picture} and Minimum thresholding, as this suited our purpose more accurately. In the third level, these binary images are smoothed and de-noised using Gaussian blur (with $3.5$ sigma) and image mean subtraction techniques, thus making them suitable for highlighting the graphemes precisely. In the next level, all the same-valued pixels that are adjacent to each other in the resulting image are grouped together to constitute a single blob, for which the minimal rectangle bounding the blobs are computed. In the penultimate level, the thus obtained bounding blob rectangles are grouped and refined by passing through the three sub-levels borrowed from the Region Proposal stage, to generate the individual wholesome symbol regions. The three sub-levels borrowed in specific from the Region Grouping sub-stage, are namely ``contained boxes removal", ``draw super box" and ``draw extended super box", with minor parametric changes. To be specific, the overlap threshold in the ``draw super box" method was dropped down to 15\% and the thresholding was removed from the ``draw extended super box" method. Finally, the resultant symbol-wise regions are used to effectively crop out the symbols from the input text-only regions. Fig~\ref{symbol_segmentation} illustrates a sample flow through this stage of the pipeline.

\subsection*{Symbol Identification}
This is the final stage of the pipeline, it takes the individually cropped images of graphemes identified from the previous stage as input and should ideally label (classify) those symbols into one of the 417 classes (the known number of Indus graphemes \cite{mahadevan1977indus}) according to the Mahadevan corpus (M77). However, in our current work, we restrict ourselves to just identifying the presence or absence of the most frequent grapheme \cite{yadav2010statistical}, the ``jar" in the input images.
We used the ``Jar-NoJar Dataset" (see Materials and Methods section) as the base dataset for this purpose and developed various image augmentation techniques inspired from Keras \cite{keras_chollet2015} to increase the size of the dataset. The image augmentation techniques used include, vertical and horizontal flips, random shear, crop, swirl, rotate, scale, and translate; with randomized parameters. This kind of augmentation is in preparation for the future goal of building a more robust and generalized model for completely classifying all the 417 graphemes of the Mahadevan corpus.

\begin{figure}[!ht]
\centering
\includegraphics[width=\textwidth]{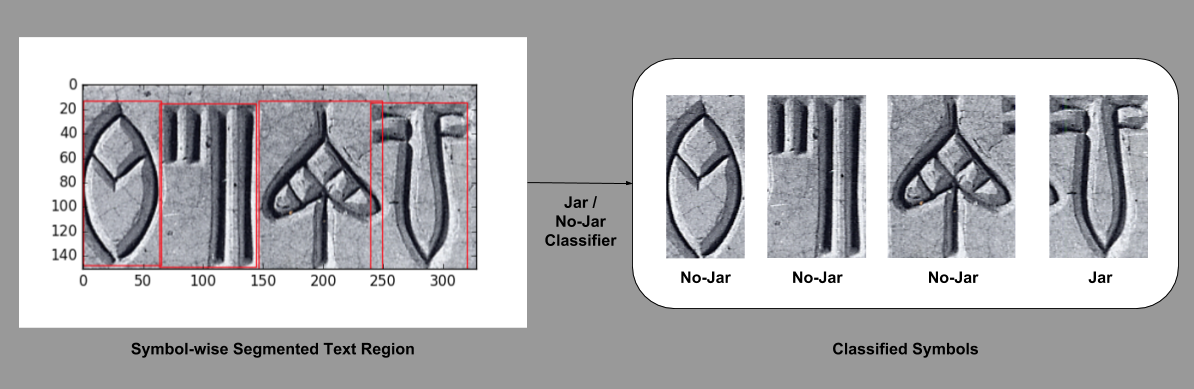}
\caption{{\bf The Symbol Identification stage}}
\label{symbol_identification}
\end{figure}

A CNN based deep-learned model was trained to perform this binary classification task, wherein any given image gets classified as either containing or not containing a ``jar" in it. A CNN architecture primarily used for optical character recognition over English text by \cite{ocr_caffe_pannous2015}, was used as the base architecture for this use case. The architecture consists of two levels of convolution layer followed by two fully-connected layers. It has a non-linearity introduced between the fully-connected layers and a dropout layer between the convolutions and full connections to reduce the chances of over-fitting the relatively small training data. The final fully-connected layer is configured to output just two values which were classified by the SoftMax classifier into the most appropriate of the two candidate classes. Fig~\ref{symbol_identification} illustrates a sample flow through this stage of the pipeline, based on a classifier capable of identifying the presence or absence of the ``jar" sign.

\subsection*{Empirical Analysis}

In order to evaluate the performance of our Indus script OCR deep learning pipeline, we isolated a set of 50 images from the original ``Indus Seals Dataset" (see Materials and Methods) gathered from Roja Muthiah Research Library (RMRL) \cite{rmrl}, which was not used to train or optimize any stage of the pipeline. The various stages in the pipeline were evaluated over these set of 50 images. Firstly, these images were fed into the Region Proposal stage of the pipeline. The Extract Seal sub-stage uses basic image processing techniques and fairly performs perfectly in all cases. Once the seal has cropped from the input images they were scaled to a standard $512x512$ or $256x256$ or left with un-scaled, depending on whichever scale gave the best results down the pipeline. The next Selective Search sub-stage produces region proposals that were classified into three classes namely ``text", ``no-text" or ``both" by the Region Classification sub-stage. The Region Classification sub-stage constituted a GoogLeNet based CNN with a classification accuracy of 89.3\% when evaluated over the validation set of the ``Text-NoText Dataset" (see Materials and Methods). Table~\ref{googlenet_table} shows the accuracy scores achieved by this classifier at various levels of the GoogLeNet CNN's network architecture.

\begin{table}[!ht]
\centering
\caption{
{\bf The Region Classification sub-stage's Text/No-Text CNN's Top-1 Accuracy Scores}}
\label{googlenet_table}
\begin{tabular}{|c|c|}
\hline
\textbf{GoogLeNet's Levels}   & \textbf{Top1 Accuracy Scores} \\ \thickhline
Level 1 ($1/3$\textsuperscript{rd} network depth) & 87.14\%                       \\ \hline
Level 2 ($2/3$\textsuperscript{rd} network depth) & 87.86\%                       \\ \hline
Level 3 (full network depth)  & 89.30\%                       \\ \hline
\end{tabular}
\end{table}

Then the Text Region Formulation sub-stage works on generating perfect text-only regions from the classified region proposals. And finally, out of the 50 test samples, complete text regions were successfully extracted out from 43 seal images, while the pipeline failed in extracting the full-text regions in the other 7 cases. However, it was successful in pulling out partial text regions that missed just a couple of symbols at max in those 7 cases too. Thus projecting a perfect case accuracy of 86\%. These text-only regions obtained, were then passed through the Symbol Segmentation stage of the pipeline to clip out the individual symbol-only regions. Eventually, of the 50 test images, after the extraction of text-only regions, the Symbol Segmentation stage succeeded in cropping out the precise symbols from 34 of them (consisting of 29 full-text regions and 5 partial text regions). While the pipeline also did fairly well in 13 other cases (consisting of 11 full-text regions and 2 partial text regions), where it was able to mark out most of the symbols individually except for a couple or more of them getting marked in entirety as single symbols, instead of separate ones. However, the remaining 3 samples failed completely in getting even a single symbol out of the proposed text region, this failure is mainly due to heavily damaged seals and a complex seal structure. Therefore, this stage's output when scored over only the perfect results gets an accuracy score of 68\% and if we do consider cases of fair performance, the accuracy score bumps up to 94\%. Table~\ref{results_table} summarizes the results of the analysis.

{
\renewcommand\extrarowheight{10pt}
\begin{table}[!ht]
\centering
\caption{
{\bf The Indus script deep learning pipeline's empirical analysis results on 50 sampled test seal images}}
\label{results_table}
\scalebox{0.76}{
\begin{tabularx}{1.34\linewidth}{|p{2.5cm}|Y|Y|Y|Y|Y|Y|Y|}
\hline
\centering\textbf{Stages \linebreak in Pipeline}                                          & \multicolumn{6}{c|}{\textbf{Output Classes}}                                                                                                                          & \textbf{Indicative Accuracies (completely perfect cases only)} \\ \thickhline
\multirow{2}{2.4cm}{\centering\textbf{Region Proposal \linebreak and \linebreak Text Region Extraction}}\bigstrut & \multicolumn{3}{c|}{\textbf{Full Text regions}}                                   & \multicolumn{3}{c|}{\textbf{Partial Text regions}}                                & \textbf{Full Text Regions} (43/50)                                      \\ \cline{2-8}
                                                                     & \multicolumn{3}{c|}{43}                                                           & \multicolumn{3}{c|}{7}                                                            & 86\%                                                                    \\ \hline
\multirow{2}{2.4cm}{\centering\textbf{Symbol Segmentation}}\bigstrut                        & \textbf{Full \linebreak Symbols} & \textbf{Partial/ \linebreak Combined Symbols} & \textbf{No\linebreak Symbols} & \textbf{Full \linebreak Symbols} & \textbf{Partial/ \linebreak Combined Symbols} & \textbf{No \linebreak Symbols} & \textbf{Full Symbols} ((29+5)/50)                                       \\ \cline{2-8}
                                                                     & 29                    & 11                                  & 3                   & 5                     & 2                                   & 0                   & 68\%                                                                    \\ \hline
\end{tabularx}}
\end{table}
}

Finally, once the symbol-only crops were obtained the Symbol Identification stage was responsible classifying symbols into those with (``Jar") and without (``No-Jar") the ``jar" sign. For which they were scaled to a standard $32x32$ size and fed into the SymbolNet CNN (see Materials Methods), which is the Jar/No-Jar Classifier. This classifier gave an accuracy score of 92.07\% when evaluated over the validation set of the ``Jar-NoJar Dataset" (see Materials and Methods).

\subsection*{Discussion}
\begin{figure}[!ht]
\centering
\includegraphics[width=\textwidth]{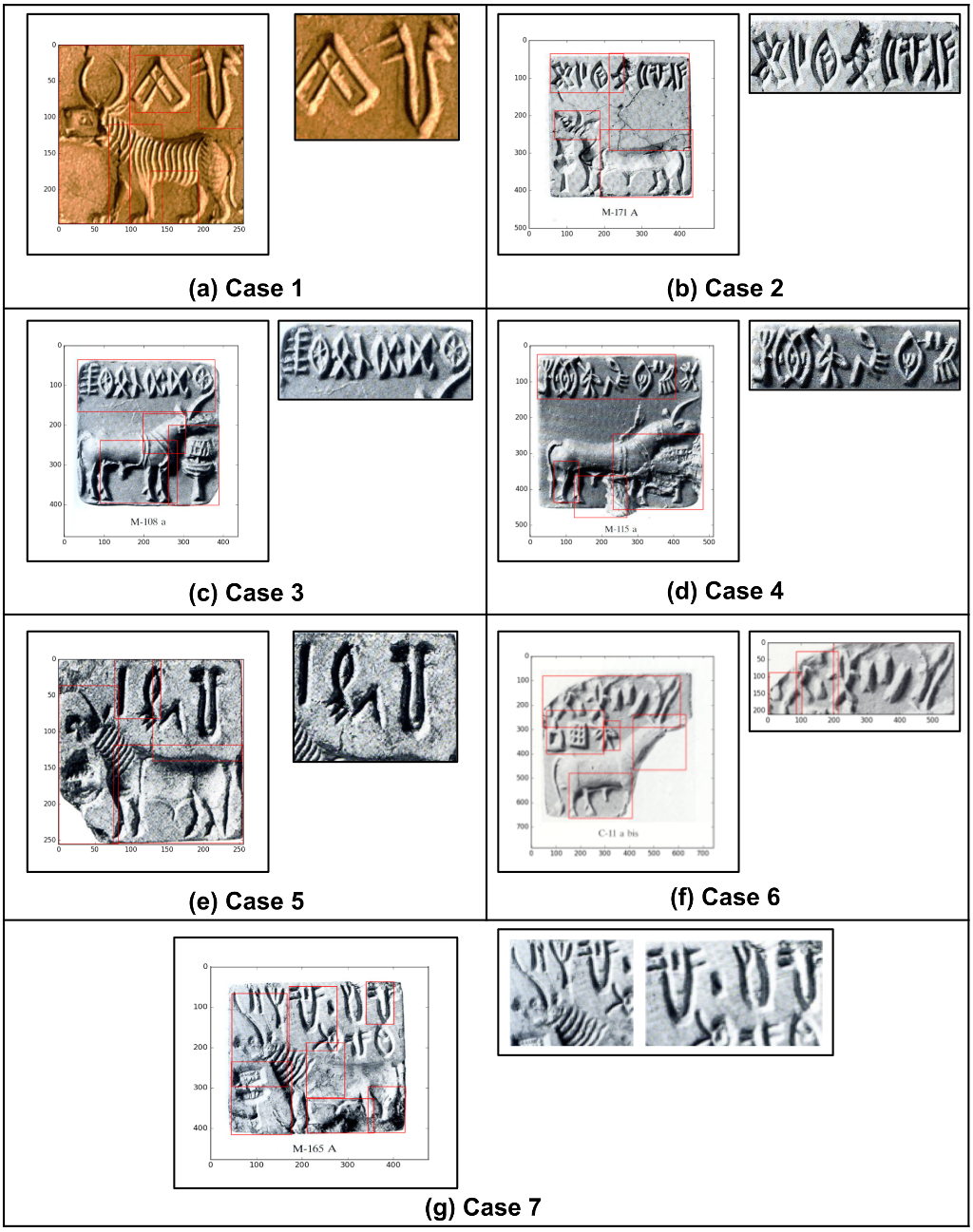}
\caption{{\bf Samples illustrating all the seven cases under the deep learning pipeline's performance discussion}}
\label{discussion_cases}
\end{figure}

There are seven cases that challenge the performance of this deep learning pipeline, the effectiveness of the system in all these cases has been discussed here. \textbf{Case 1}: The generality of the model; it not only works well with perfectly scanned or photographed seal images, it work fine with noisy images as well, as it was also trained with images crawled from the web, which exposed the pipeline to casually photographed scenarios. \textbf{Case 2}: Straightforward seals with a simple structure; without much clutter and damage the model was perfect in extracting the individual symbols out of the seals as the images, in this case, were noise free. \textbf{Case 3}: Simple seals with unseen symbols in them; the model's Text/No-Text classifier of the Region Classification sub-stage has learned to generalize over the dataset in hand, thanks to the deep features learned by the model, it had no problems in recognizing unseen symbols/text parts as the features used to classify the regions as text/no-text/both were not over-fitting the dataset in hand. \textbf{Case 4}: Incomplete region proposals; in some cases the Selective Search sub-module of the region proposal stage might fail to recognize all the objects (symbols, animals, and humans), but in most cases, the hierarchical region grouping mechanisms cover up for this. However, at maximum, a couple of symbols get missed out in some exceptional cases. In this specific case illustrated in Fig~\ref{discussion_cases}d, the crab sign has been missed from the proposed text region bounding box. \textbf{Case 5}: Seal images with background noise; the pipeline developed is not tolerant to the seal being surrounded by some background, but this case is successfully handled by the Extract Seal sub-stage of the Region Proposal stage, which uses a modified Canny Edge detection \cite{canny1986computational} based algorithm to cut off the given seal image only from its background noise. \textbf{Case 6}: Heavily damaged and worn out seals; the pipeline succeeds in extracting the text regions out of the Indus seals, but in this case fails to extract the symbols perfectly out of these text regions, as the damage in the seals affects one or more continuous symbols, making the system to combine them into one, rather than splitting them individually. \textbf{Case 7}: Completely disorganized seals; such seals don't have any seal structure, they have a haphazard arrangement of symbols in the seal as in Fig~\ref{discussion_cases}g. Our pipeline still does a decent job in getting most of the text regions out of it, missing a few owing to a more convoluted seal structure. Fig~\ref{discussion_cases} exhibits samples of all the cases discussed so far.

\section*{Materials and Methods}

\subsubsection*{Datasets}
Building a deep learning pipeline to effectively perform OCR on Indus script inscriptions requires an image corpus of the Indus artifacts. The base dataset of images, either scans or photographs of Indus artifacts that were used to formulate the other datasets was very limited in size and variety. They were mainly sourced from Roja Muthiah Research Library (RMRL) \cite{rmrl} along with their corresponding text mappings according to Mahadevan's corpus \cite{mahadevan1977indus} and by scraping the web with the Google Image Search API \cite{google-search-api}, especially from the harappa.com \cite{harappawebsite}. This together constitutes the \textbf{Indus Seals Dataset}. It includes 800 high-resolution scanned images of Indus artifacts indexed from CISI M0101 to CISI M0620 at RMRL and 350 images of Indus artifacts from the web. When sourcing the data from the web, for a given search term the number of results were restricted to top 100, with an interest in limiting the irrelevant images being pulled from the web. Further, the considerable amount of noise in the crawled images had to be filtered manually, resulting in 350 images from the total 1000 images retrieved. As a whole, this base dataset comprised of 1150 images of Indus seals. Figures in \nameref{S1_Fig} and \nameref{S2_Fig} shows samples from the RMRL and crawled versions of the Indus Seals Dataset, respectively. The different variations of this corpus were used in various stages of the pipeline.

The \textbf{Text-NoText Dataset} used by the Region Classification level of Text Region Extraction stage, consisted of three classes namely, ``text", ``no-text" and ``both". The candidates for this dataset were obtained by applying the vanilla version of Selective Search \cite{uijlings2013selective} algorithm over the images of the base dataset. The resulting regions of interest were manually filtered to remove noisy images and those region images containing, not containing and partly containing the Indus text/symbols were classified into the ``text", ``no-text" and ``both" classes, respectively. In total, this dataset consisted of 2091 entries of which 652 were ``text", 1055 were ``no-text" and 384 were ``both" regions. The dataset was split in a 70:30 ratio for training and validation purposes respectively, following the stratified random sampling technique, wherein the data proportions across classes are equally weighted during the split. In summary, of the total 2091 entries, 1466 regions (``text" - 458, ``no-text" - 739, ``both" - 269) were used for training and the remaining 625 regions (``text" - 194, ``no-text" - 316, ``both" - 115) were used for validation. Figure in \nameref{S3_Fig} shows sample entries from the Text-NoText Dataset.

The \textbf{Jar-NoJar Dataset} employed in the Symbol Identification stage consisted of two classes namely, the ``Jar" and ``NoJar". It was formulated from the crawled version of the ``Indus Seals Dataset" containing 350 Indus artifact images. These images were manually classified into two classes, based on the presence or absence of the ``jar" sign. The resulting dataset has 168 images with a ``jar" sign and 182 without a ``Jar" sign, of the 350 images in total. This dataset was split in a 70-30 stratified ratio, with 246 images (``Jar" - 118, ``NoJar" - 128) for training and 104 images (``Jar" - 50, ``NoJar" - 54) for validation. Figure in \nameref{S4_Fig} shows sample entries from the Jar-NoJar Dataset.

\subsubsection*{Canny Edge Detection}
The Canny edge detection \cite{canny1986computational} is a multi-stage algorithm with two significant features namely, Non-Maximum Suppression and the Hysteresis Process, wherein the edges' candidates which are not dominant in their neighborhood aren't considered to be edges and given a candidate is in the neighborhood of an edge the threshold is lowered, while moving along the candidates. These traits best suits the Indus artifacts use case in hand. In order to further optimize this for the purpose of seal extraction from given images, the lower and upper thresholds of the algorithm are adjusted according to the image at hand, by computing the median of the single channel pixel intensities.

\subsubsection*{Selective Search Algorithm}
It is a region proposal algorithm \cite{uijlings2013selective} used in the Selective Search level of the Region Proposal stage of the pipeline. It combines the advantages of exhaustive search and segmentation. Like segmentation, the image structure is used to guide sampling and like an exhaustive search, all possible object locations are captured invariant of size and scale, making it an optimal choice for our case. Under the hood, it performs hierarchical grouping of the region proposals based on color, texture, size and fills to propose the best candidate regions. It was also used as the Region Proposal mechanism for Regions with CNN \cite{rcnn_girshick2014rich}, proving its compatibility with deep architectures. Further to suit our needs, the algorithm's four free parameters were fine-tuned by manual grid search, and finalized on: ``Scale" - 350, 450, 500 (higher the value larger the clusters in Felzenszwalb segmentation \cite{felzenszwalb2004efficient}), ``Sigma" - 0.8 (width of Gaussian kernel for Felzenszwalb segmentation \cite{felzenszwalb2004efficient}), ``Minimum Size" - 30, 60, 120 (Minimum component size for Felzenszwalb segmentation \cite{felzenszwalb2004efficient}), ``Minimum Area" - 2000 (Minimum area of a region proposed). Figure in \nameref{S5_Fig} shows a case where the modified Selective Search was applied to an Indus seal image.

\subsection*{Convolutional Neural Networks}
The Convolution Neural Networks (CNN) \cite{cnn_lenet_lecun1998gradient} are powerful deep learning algorithms that operate explicitly on images. They are inspired by the actual working of the visual cortex of the animal eye and how the neurons process images and respond towards overlapping regions tiling the visual field. Unlike other feed-forward artificial neural networks, the CNNs have different types of layers, the convolution layer, pooling layer, LRN \cite{alexnet_lrn_relu}, ReLU \cite{alexnet_lrn_relu}, dropout layer \cite{dropout_hinton2012improving}, fully-connected layer and SoftMax layer, to name a few. The blueprint that describes how these layers are stacked together along with the hyperparameter configurations is referred to as the CNN's Architecture. Both the CNNs described in our work here were built, trained, transfer learned and fine-tuned using the ``Caffe" deep learning framework developed by “Yangqing Jia” at the Berkeley Vision and Learning Center \cite{jia2014caffe}. It is primarily in C++ with Python bindings and it uses the Google's Protocol Buffers specification to specify the CNN architectures as text configuration files, that are translated to in-memory CNN architectures by the framework. Its specific implementation with cuDNN trained on Nvidia GPUs was 40 times faster than basic CPU based training. Hence, all of our models were trained on a Nvidia GTX 660M graphics card with 2GB memory. Two important stages of our proposed pipeline use CNNs namely, ``Region Classification" sub-stage and ``Symbol Identification" stage.

\subsubsection*{Region Classification}
The Region Classification sub-stage of the Text Region Extraction stage in the pipeline employs the \textbf{GoogLeNet} \cite{googlenet_szegedy2015going} CNN architecture for classifying the input regions into ``text", ``no-text" and ``both". We also surveyed other famous CNN architectures \cite{CS231n} such as LeNet \cite{cnn_lenet_lecun1998gradient}, AlexNet \cite{alexnet_lrn_relu}, and VGGNet \cite{vgg_simonyan2014very}, designed for image classification problems, but GoogLeNet was the most deepest, computationally efficient and lightweight architecture that delivered state-of-the-art results in the ImageNet Large-Scale Visual Recognition Challenge 2014 \cite{ImageNet_russakovsky2015imagenet} (ILSVRC14). In specific, it harnesses the Network-In-Network architecture \cite{NIN_lin2013network} for improving the representational power of a neural network. This 22 layers deep network has 9 characteristic Inception networks stacked over each other. They are grouped into 3 levels and the predictions happen at each level along the network and it was observed that the deepest level gives the best accuracies. The first level has two inception layers (3a and 3b), followed by level two with 5 inception layers (4a, 4b, 4c, 4d, and 4e) and finally level three with two layers (5a and 5b). Figure in \nameref{S6_Fig} shows the overall GoogLeNet CNN architecture employed in this stage and the figure in \nameref{S7_Fig} delineates the NIN inception module of the architecture, in specific.

The ``Text-NoText Dataset" has just a few thousand data points, which engenders the inability to experiment with new CNN architectures and train models from scratch. Therefore, we opted for the practice of \textbf{Transfer Learning} \cite{yosinski2014transferable}, where the weights and rich filters learned by the CNN on some other larger and generic dataset are fine tuned and transfer learned to suit our primary dataset. In our case, we used the GoogLeNet's ImageNet trained final snapshot released by Berkeley Vision and Learning Center (BVLC) \cite{bvlc_googlenet} to initialize our CNN model. Based on the observations that the earlier layers of a CNN describe more generic features like Gabor filters, blob detectors or edge detectors and as the network progresses the features learned in the later layers of the CNN become more specific to the dataset being trained on. Therefore, we tweaked the GoogLeNet \cite{googlenet_szegedy2015going} architecture's parameters to disable learning until the first 6 inception layers (3a, 3b, 4a, 4b, 4c, and 4d) by setting their learning rates to 0, thus preserving the rich lower level ImageNet features. And, the last learning rates of the last three inception layers (4e, 5a, 5b) were stepped up by 2 times from the default, to enable active learning. In addition to this, the input Indus seal images were randomly cropped to $224x224$ and the default GoogLeNet's mean values (R-104, G-117, B-123) were used for normalizing them.

\begin{figure}[!ht]
\centering
\includegraphics[width=1.1\textwidth]{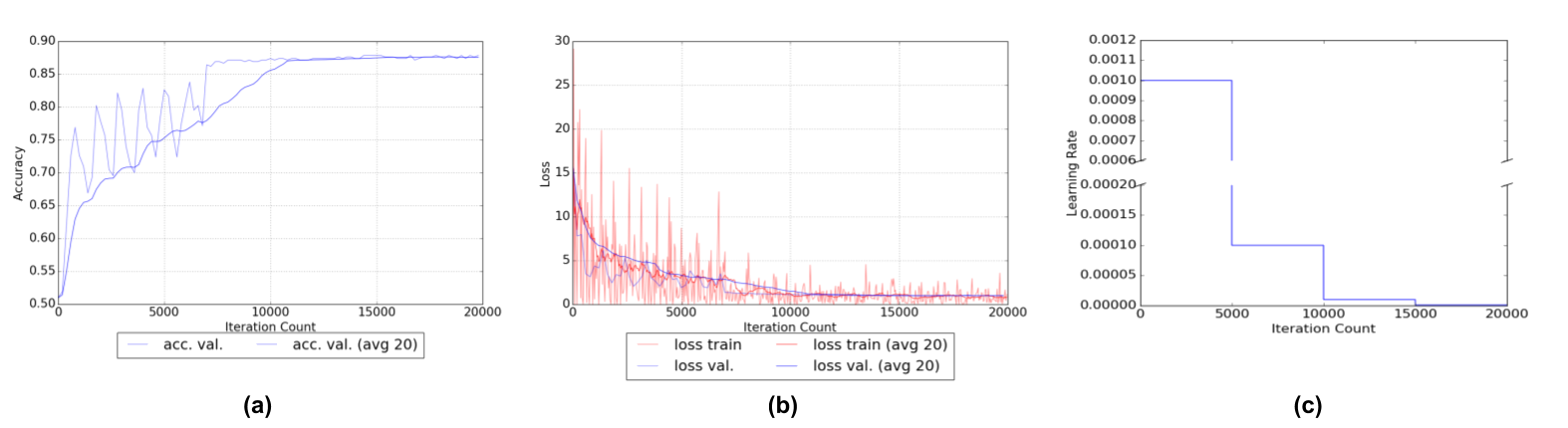}
\caption{{\bf The Region Classification CNN's: (a): Accuracy plot, (b): Loss plot, and (c): Learning Rate plot}}
\label{googlenet_loss_accu_lr}
\end{figure}

The Stochastic Gradient Descent (SGD) solver was used for training the GoogLeNet model. The solver was configured with a low base learning rate of 0.001, to avoid greatly offsetting the already learned weights from ImageNet in the last three Inception layers. The ``step" learning rate update/decay policy Eq~(\ref{eq:stepLR}) was used, and setup with a step size of 1000, momentum of 0.9, weight decay rate of 0.0002 and gamma value of 0.1. The training and validation batch sizes were 30 and 20, with a maximum iteration of 20000 (approximately 14 epochs). Fig~\ref{googlenet_loss_accu_lr} shows a graphical representation of the progressions in the accuracy, loss and learning rate over the number of iterations elapsed in training the Text/No-Text Region Classifier.

\begin{eqnarray}
\label{eq:stepLR}
  \mathrm{newLearningRate} = \mathbf{baseLearningRate} * \mathbf{gamma}^{(floor(\mathbf{iterationNumber} / \mathbf{step}))}
\end{eqnarray}

\subsubsection*{Symbol Identification}
\begin{figure}[!ht]
\centering
\includegraphics[width=1.08\textwidth]{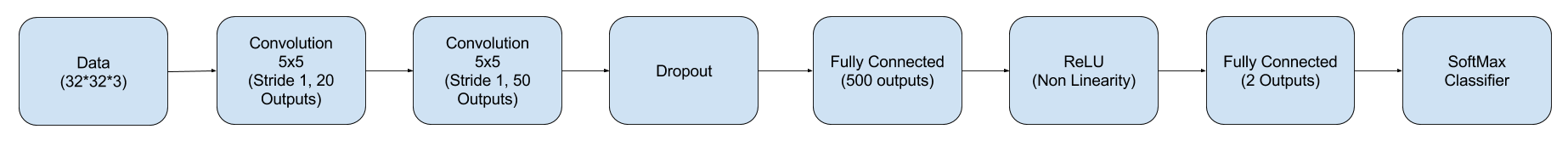}
\caption{{\bf The SymbolNet CNN Architecture \cite{ocr_caffe_pannous2015}}}
\label{symbolnet}
\end{figure}

The Symbol Identification stage in the pipeline is yet another place where a deep learning classifier was trained to identify the presence or absence the ``jar" sign in the input symbol-only regions. The ``Jar-NoJar Dataset" was used for this purpose, the images were resized to $32x32$, converted to grayscale, and the pixel intensities were normalized over 255 as a part of the data preparation. The CNN architecture used in building this stage was inspired from \cite{ocr_caffe_pannous2015}. This architecture has two Convolution layers stacked over each other, both with a kernel size of $5X5$ and a stride shift of 1, and configured to produce 20 and 50 output filter banks, respectively. This layer is followed by a Dropout \cite{dropout_hinton2012improving} layer, to prevent from over-fitting the small dataset in hand. This is followed by a ReLU layer, for non-linear rectification sandwiched in between two fully-connected layers of 500 and 2 outputs, respectively. The final fully-connected layer's two output values were classified by the SoftMax classifier into the most appropriate of the two candidate classes. We call this architecture the SymbolNet. Fig~\ref{symbolnet} shows the overall SymbolNet CNN architecture employed in this stage.

\begin{figure}[!ht]
\centering
\includegraphics[width=1.1\textwidth]{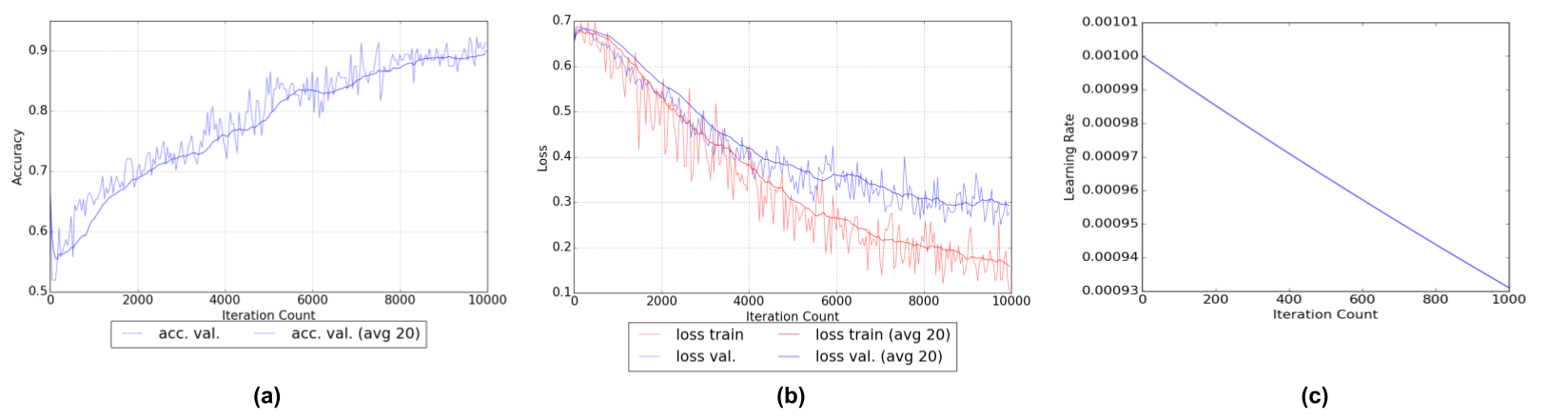}
\caption{{\bf The Symbol Identification CNN's: (a): Accuracy plot, (b): Loss plot, and (c): Learning Rate plot}}
\label{symbolnet_loss_accu_lr}
\end{figure}

The Stochastic Gradient Descent (SGD) solver was used for training the SymbolNet model from scratch without any pre-trained weight initialization. The SGD solver was configured with a base learning rate of 0.001, a momentum of 0.9, and a weight decay of 0.0005. The ``inv" learning rate update/decay policy Eq~(\ref{eq:invLR}) was used, and setup with a gamma and power value of 0.0001 and 0.75 respectively. The training and validation batch sizes were 100 and 52, with a maximum iteration of 10000. The training was CPU based. Fig~\ref{symbolnet_loss_accu_lr} shows a graphical representation of the progressions in the accuracy, loss and learning rate over the number of iterations elapsed in training the Jar/No-Jar Symbol Classifier.

\begin{eqnarray}
\label{eq:invLR}
  \mathrm{newLearningRate} = \mathbf{baseLearningRate} * (1 + \mathbf{gamma} * \mathbf{iterationNumber})^{- power}
\end{eqnarray}

\section*{Conclusion}
In conclusion, the proposed deep learning pipeline's architecture is capable of reducing a complex optical character recognition problem on the Indus Script into a relatively simpler image classification problem and further detecting the presence and absence of the ``jar" sign. It performs perfectly well in almost all cases, but in some rare cases, the results are not precise when the seals are heavily damaged or the text arrangement in the seal structure is too complex for the pipeline to recognize. From all the discussions it is clear that the Text/No-Text classifier in the Region Classification sub-stage is very mature and performs well, thus contributing to a more stable pipeline, even if some cases get missed by this module, they are covered up by the handcrafted region proposal mechanisms devised as a part of this pipeline. However, some rare limitations still exist in the pipeline, such as in Case 4, due to the Selective Search misbehavior that fails to recognize certain symbols along the seal edges. And in cases 6 and 7, as they dealt with damages in the seal, but the pipeline has done a decent job in getting at least the partial regions out of it. And as a forerunner to a full-fledged symbol identification stage that follows the symbol segmentation stage, the experimentation with the ``jar" sign has proved the feasibility of the symbol classification task. This work also lays a foundation for building a real-time system that can ``read'' the Indus script from images captured on a camera-enabled mobile computing device. The implementation of a mobile application is part of ongoing work.

Our work here describes the deep-learned pipeline architecture for optical character recognition with special emphasis on the Indus script. However, the architecture proposed here can be seamlessly trained over a completely new set of ancient inscriptions other than Indus script, given the availability of a sufficient number of images and an existing corpus. Every stage of our pipelined architecture is independent of dataset-specific approaches, it embraces all the challenges faced in visually recognizing any ancient inscription in general, thus engendering a flexible optical character recognition engine. Especially, the core phases in the pipeline are completely based on deep learning techniques that have proved to be invariant to the various common computer vision challenges, thus making them adaptive to any dataset in hand. The major difficulty faced when trying to solve any problem in the field of epigraphical research, is due to the lack of or the meagre availability of a standardized corpus of the inscriptions under study. With this generic deep-learned OCR pipeline, our work here can contribute to substantially reducing the tedium and human effort required in preparing epigraphic corpora from archaeological artifacts. It also demonstrates the promise of deep neural networks in the epigraphic research with the tantalizing possibility of providing a more expressive representation for the automated recognition of syntax. The significance of the latter for decipherment efforts does not need further comment. We conclude with the hope that deep learning will be applied in many other areas of the digital humanities which involve pattern recognition and classification in high-dimensional data spaces.

\section*{Supporting Information}

\paragraph*{S1 Fig.}
\label{S1_Fig}
{\bf A snapshot of the ``Indus Seals Dataset" (RMRL)}

\paragraph*{S2 Fig.}
\label{S2_Fig}
{\bf A snapshot of the ``Indus Seals Dataset" (Crawled)}

\paragraph*{S3 Fig.}
\label{S3_Fig}
{\bf A snapshot of the ``Text-NoText Dataset"}

\paragraph*{S4 Fig.}
\label{S4_Fig}
{\bf A snapshot of the ``Jar-NoJar Dataset"}

\paragraph*{S5 Fig.}
\label{S5_Fig}
{\bf Selective Search applied to an Indus seal image}

\paragraph*{S6 Fig.}
\label{S6_Fig}
{\bf The GoogLeNet CNN Architecture \cite{googlenet_szegedy2015going}}

\paragraph*{S7 Fig.}
\label{S7_Fig}
{\bf The Inception Module in GoogLeNet \cite{googlenet_szegedy2015going}}

\section*{Acknowledgments}
The second author's research is partially supported by a Google Faculty Research Award on ``Machine learning of syntax in undeciphered scripts". We acknowledge the generous assistance of Suresh Babu of the Indus Research Centre and G. Sundar of the Roja Muthiah Research Library in providing images of the Indus seals. We thank Harappa.com \cite{harappawebsite} for their kind permission to use an image of the Indus seal in this paper. We record our gratitude to Iravatham Mahadevan for being a continued source of wisdom and inspiration in research on the Indus civilization.


\begin{thebibliography}{35}

\bibitem{knorozov1968proto}
Knorozov Y, Volchok B, Gurov N. Proto-Indica: Brief Report on the Investigation
  of the Proto-Indian Texts; 1968.

\bibitem{koskenniemi1970method}
Koskenniemi S, Parpola A, Parpola S.
\newblock A method to classify characters of unknown ancient scripts.
\newblock Linguistics. 1970;8(61):65--91.

\bibitem{koskenniemi1981syntactic}
Koskenniemi K.
\newblock Syntactic methods in the study of the Indus script.
\newblock Studia Orientalia Electronica. 1981;50:125--136.

\bibitem{rao2009entropic}
Rao RP, Yadav N, Vahia MN, Joglekar H, Adhikari R, Mahadevan I.
\newblock Entropic evidence for linguistic structure in the Indus script.
\newblock Science. 2009;324(5931):1165.

\bibitem{rao2009markov}
Rao RP, Yadav N, Vahia MN, Joglekar H, Adhikari R, Mahadevan I.
\newblock A Markov model of the Indus script.
\newblock Proceedings of the National Academy of Sciences.
  2009;106(33):13685--13690.

\bibitem{deep_learning2015}
LeCun Y, Bengio Y, Hinton G.
\newblock Deep learning.
\newblock Nature. 2015;521:436–444.

\bibitem{deep_learning_conspiracy2015}
Schmidhuber J. Deep Learning Conspiracy; 2015.
\newblock Available from:
  \url{https://plus.google.com/100849856540000067209/posts/9BDtGwCDL7D}.

\bibitem{kenoyer1998ancient}
Kenoyer JM.
\newblock Ancient Cities of the Indus Valley Civilization.
\newblock American Institute of Pakistan studies. Oxford University Press;
  1998.
\newblock Available from:
  \url{https://books.google.co.in/books?id=DK3tAAAAMAAJ}.

\bibitem{cnn_lenet_lecun1998gradient}
LeCun Y, Bottou L, Bengio Y, Haffner P.
\newblock Gradient-based learning applied to document recognition.
\newblock Proceedings of the IEEE. 1998;86(11):2278--2324.

\bibitem{mahadevan1977indus}
Mahadevan I.
\newblock The Indus script: texts, concordance, and tables.
\newblock Memoirs of the Archaeological Survey of India. Archaeological Survey
  of India; 1977.
\newblock Available from:
  \url{https://books.google.co.in/books?id=7WkMAQAAMAAJ}.

\bibitem{ocr_indus_seals2016}
Palaniappan S, Adhikari R. OCR on Indus Seals; 2016.
\newblock Available from:
  \url{https://github.com/tpsatish95/OCR-on-Indus-Seals}.

\bibitem{canny1986computational}
Canny J.
\newblock A computational approach to edge detection.
\newblock IEEE Transactions on pattern analysis and machine intelligence.
  1986;(6):679--698.

\bibitem{uijlings2013selective}
Uijlings JR, van~de Sande KE, Gevers T, Smeulders AW.
\newblock Selective search for object recognition.
\newblock International journal of computer vision. 2013;104(2):154--171.

\bibitem{ImageNet_russakovsky2015imagenet}
Russakovsky O, Deng J, Su H, Krause J, Satheesh S, Ma S, et~al.
\newblock Imagenet large scale visual recognition challenge.
\newblock International Journal of Computer Vision. 2015;115(3):211--252.

\bibitem{otsu1975threshold}
Otsu N.
\newblock A threshold selection method from gray-level histograms.
\newblock Automatica. 1975;11(285-296):23--27.

\bibitem{li1993minimum}
Li CH, Lee C.
\newblock Minimum cross entropy thresholding.
\newblock Pattern Recognition. 1993;26(4):617--625.

\bibitem{yen1995new}
Yen JC, Chang FJ, Chang S.
\newblock A new criterion for automatic multilevel thresholding.
\newblock IEEE Transactions on Image Processing. 1995;4(3):370--378.

\bibitem{isodata_ridler1978picture}
Ridler T, Calvard S.
\newblock Picture thresholding using an iterative selection method.
\newblock IEEE trans syst Man Cybern. 1978;8(8):630--632.

\bibitem{yadav2010statistical}
Yadav N, Joglekar H, Rao RP, Vahia MN, Adhikari R, Mahadevan I.
\newblock Statistical analysis of the Indus script using n-grams.
\newblock PLoS One. 2010;5(3):e9506.

\bibitem{keras_chollet2015}
Chollet F. Keras; 2015.
\newblock Available from: \url{https://github.com/fchollet/keras}.

\bibitem{ocr_caffe_pannous2015}
Pannous. OCR using Caffe; 2015.
\newblock Available from: \url{https://github.com/pannous/caffe-ocr}.

\bibitem{rmrl}
Roja Muthiah Research Library;.
\newblock Available from: \url{http://www.rmrl.in/}.

\bibitem{google-search-api}
Custom Search  |  Google Developers;.
\newblock Available from: \url{https://developers.google.com/custom-search/}.

\bibitem{harappawebsite}
Harappa | The Ancient Indus Civilization;.
\newblock Available from: \url{https://www.harappa.com/}.

\bibitem{rcnn_girshick2014rich}
Girshick R, Donahue J, Darrell T, Malik J.
\newblock Rich feature hierarchies for accurate object detection and semantic
  segmentation.
\newblock In: Proceedings of the IEEE conference on computer vision and pattern
  recognition; 2014. p. 580--587.

\bibitem{felzenszwalb2004efficient}
Felzenszwalb PF, Huttenlocher DP.
\newblock Efficient graph-based image segmentation.
\newblock International Journal of Computer Vision. 2004;59(2):167--181.

\bibitem{alexnet_lrn_relu}
Krizhevsky A, Sutskever I, Hinton GE.
\newblock Imagenet classification with deep convolutional neural networks.
\newblock In: Advances in neural information processing systems; 2012. p.
  1097--1105.

\bibitem{dropout_hinton2012improving}
Hinton GE, Srivastava N, Krizhevsky A, Sutskever I, Salakhutdinov RR.
\newblock Improving neural networks by preventing co-adaptation of feature
  detectors.
\newblock arXiv preprint arXiv:12070580. 2012;.

\bibitem{jia2014caffe}
Jia Y, Shelhamer E, Donahue J, Karayev S, Long J, Girshick R, et~al.
\newblock Caffe: Convolutional Architecture for Fast Feature Embedding.
\newblock arXiv preprint arXiv:14085093. 2014;.

\bibitem{googlenet_szegedy2015going}
Szegedy C, Liu W, Jia Y, Sermanet P, Reed S, Anguelov D, et~al.
\newblock Going deeper with convolutions.
\newblock In: Proceedings of the IEEE Conference on Computer Vision and Pattern
  Recognition; 2015. p. 1--9.

\bibitem{CS231n}
Karpathy A. CS231n Convolutional Neural Networks for Visual Recognition;.
\newblock Available from: \url{http://cs231n.github.io/}.

\bibitem{vgg_simonyan2014very}
Simonyan K, Zisserman A.
\newblock Very deep convolutional networks for large-scale image recognition.
\newblock arXiv preprint arXiv:14091556. 2014;.

\bibitem{NIN_lin2013network}
Lin M, Chen Q, Yan S.
\newblock Network in network.
\newblock arXiv preprint arXiv:13124400. 2013;.

\bibitem{yosinski2014transferable}
Yosinski J, Clune J, Bengio Y, Lipson H.
\newblock How transferable are features in deep neural networks?
\newblock In: Advances in neural information processing systems; 2014. p.
  3320--3328.

\bibitem{bvlc_googlenet}
Berkeley Vision and Learning Center. BVLC GooLeNet ILSVRC 2014 Snapshot; 2015.
\newblock Available from:
  \url{https://github.com/BVLC/caffe/tree/master/models/bvlc_googlenet}.

\end{thebibliography}
\end{document}


\begin{flushleft}
{\Large
\textbf{{Supporting Information for Deep Learning the Indus Script}}
}
\newline
\end{flushleft}

\begin{figure}[!ht]
\centering
\includegraphics[width=\textwidth]{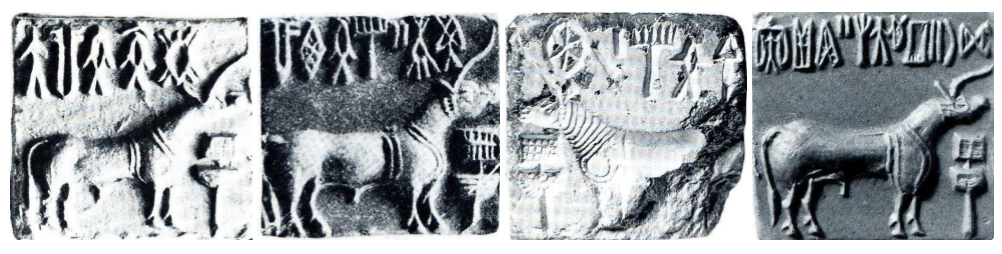}
\caption{{\bf A snapshot of the ``Indus Seals Dataset" (RMRL)}}
\label{S1_Fig}
\end{figure}

\begin{figure}[!ht]
\centering
\includegraphics[width=\textwidth]{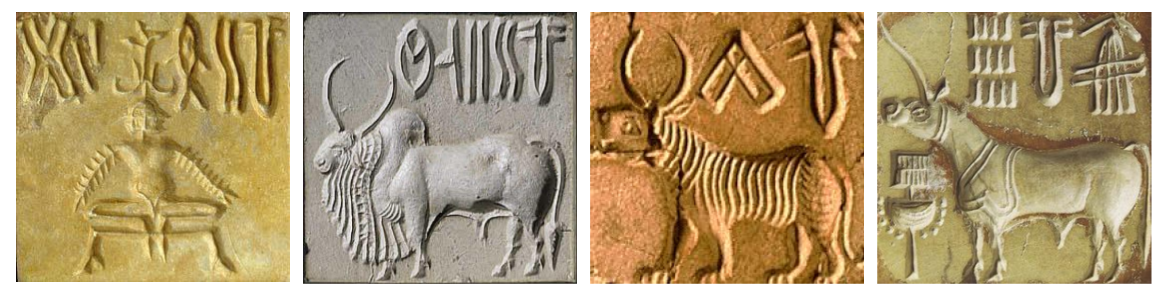}
\caption{{\bf A snapshot of the ``Indus Seals Dataset" (Crawled)}}
\label{S2_Fig}
\end{figure}

\begin{figure}[!ht]
\centering
\includegraphics[width=\textwidth]{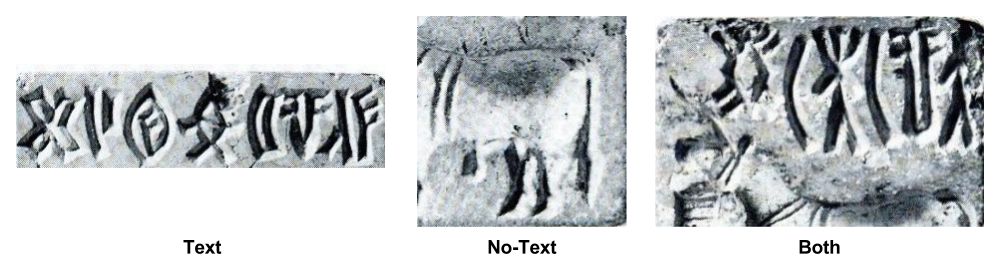}
\caption{{\bf A snapshot of the ``Text-NoText Dataset"}}
\label{S3_Fig}
\end{figure}

\begin{figure}[!ht]
\centering
\includegraphics[width=\textwidth]{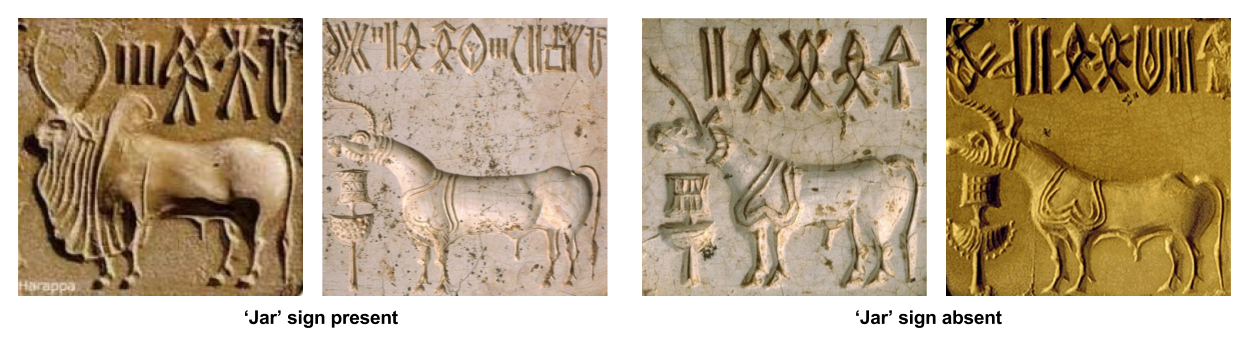}
\caption{{\bf A snapshot of the ``Jar-NoJar Dataset"}}
\label{S4_Fig}
\end{figure}

\begin{figure}[!ht]
\centering
\includegraphics[width=\textwidth]{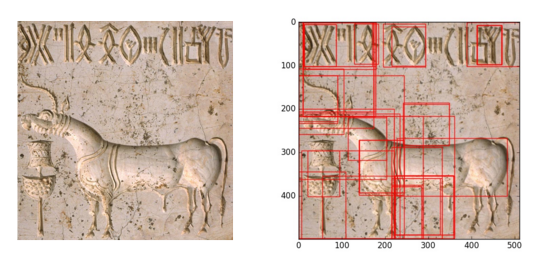}
\caption{{\bf Selective Search applied to an Indus seal image}}
\label{S5_Fig}
\end{figure}

\begin{figure}[!ht]
\centering
\includegraphics[width=\textwidth]{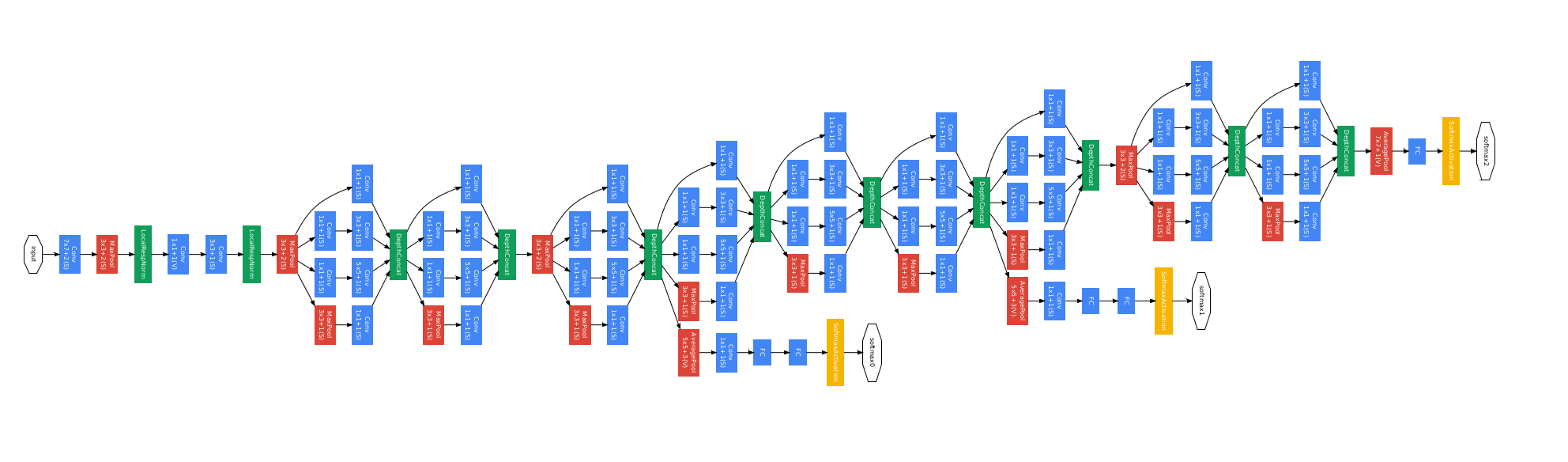}
\caption{{\bf The GoogLeNet CNN Architecture}}
\label{S6_Fig}
\end{figure}

\begin{figure}[!ht]
\centering
\includegraphics[width=\textwidth]{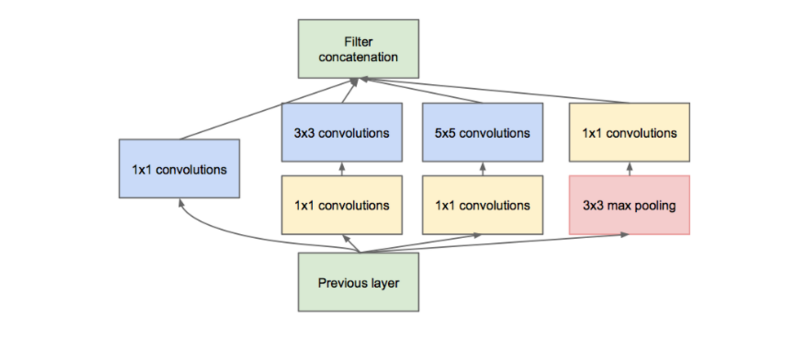}
\caption{{\bf The Inception Module in GoogLeNet}}
\label{S7_Fig}
\end{figure}